# A Security Steganography Scheme Based On HDR Image

Wei Gao
School of Information and Communication Engineering, University of Electronic Science and Technology of China, Chengdu 611731 China +86 61831246
dmxlvu@163.com

Yongqing Huo*
School of Information and Communication Engineering, University of Electronic Science and Technology of China, Chengdu 611731 China +86 61831246
hyq980132@uestc.edu.cn

Yan Qiao
School of Information and Communication Engineering, University of Electronic Science and Technology of China, Chengdu 611731 China +86 61831246
qiao790273917@163.com

## ABSTRACT
It is widely recognized that the image format is crucial to steganography for that each individual format has its unique properties. Nowadays, the most famous approach of digital image steganography is to combine a well-defined distortion function with efficient practical codes such as STC. And numerous researches are concentrated on spatial domain and jpeg domain. However, whether in spatial domain or jpeg domain, high payload (e.g., 0.5 bit per pixel) is not secure enough. In this paper, we propose a novel adaptive steganography scheme based on 32-bit HDR (High dynamic range) format and Norm IEEE 754. Experiments show that the steganographic method can achieve satisfactory security under payload from 0.3bpp to 0.5bpp.

## CCS Concepts
• **Information systems**→**Information system security**→**Image security.**

## Keywords
Steganography; security; HDR; high payload.

## 1. INTRODUCTION
Steganography is an art aim at hiding the existence itself under the attack from steganalysis [2]. Different from cryptography, once the steganography behavior is exposed, the communication lose efficacy. Therefore, steganography scheme should not draw suspicion from both visual observation and statistics. To protect visual consistency, no matter in spatial domain or jpeg domain, the sender tends to embed the secret message in the least significant bit. While to minimize statistical detectability, modern steganography can be formulated as a source coding problem that minimizes a well-defined distortion function [4]. Currently, thanks to the effective STC, designing a more suitably defined distortion function turns into the foremost task for steganographer. In this trend, there comes to born many powerful steganography techniques, including HUGO, WOW, and UNIWARD.

In recent years, investigations are highly concentrated on the spatial domain and the jpeg domain for that they are most suitable for steganography [7]. In the spatial domain, secret message bits are generally embedded in the least significant bit of the pixel values. The sufficient redundant space in the spatial domain can bear higher payload so that we can communicate more messages through spatial domain. While in the jpeg domain, the least significant bit of the nonzero DCT coefficients are usually utilized. Because of not implementing embedding operation straightly on pixel values, steganography in jpeg domain tend to be more secure. But, neither of them can ensure security and high payload at the same time which push us to do further investigation.

Currently, the vast majority of color images are represented with a byte per pixel for each of the red, green, and blue channels [1]. And modern steganography is mainly based on such 8 bit low-dynamic-range (LDR) images. However the emerging High-dynamic-range (HDR) image whose pixel is single precision floating number is seldom researched.

In this paper, 32-bit HDR format and Norm IEEE754 are employed to form a maximum of 10LSB adaptive embedding procedure. The proposed scheme is effective to different distortion measure. Experiments show that the steganography scheme can achieves good performance under payload from 0.3bpp to 0.5bpp.

The paper is organized as follows. In the next section, we introduce the notations，the fundamentally Norm IEEE754 and the concept on minimizing additive distortion. The proposed scheme is detailed in Section 3. The experiment results of the security of the proposed scheme are displayed in Section 4. And the paper is concluded in Section 5.

## 2. PRELIMINARIES
### 2.1 Notations
$X = (x_{i,j})^{n_1 \times n_2}$, $Y = (y_{i,j})^{n_1 \times n_2}$ are respectively used to denote 32-bit gray-scale cover image (of $n_1 \times n_2$ size, luminance channel only ) and corresponding stego image. The $k^{th}$ significant bit of $x_{i,j}$, $y_{i,j}$ under the norm IEEE754 are respectively denoted by $x_{i,j}^k$, $y_{i,j}^k$. The proposed scheme is restricted for the binary embedding condition for which $y_{i,j}^k \in \{x_{i,j}^k, \overline{x}_{i,j}^k\}$.

### 2.2 Norm IEEE754
The IEEE Standard for Floating-Point Arithmetic (IEEE 754) is a technical standard for floating-arithmetic established in 1985 by the Institute of Electrical and Electronics Engineers (IEEE). It is employed in our scheme to embed secret messages into single precision floating number.

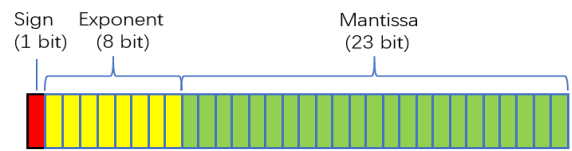

**Figure 1. Norm IEEE754 of single precision number.**

32 bits are assigned to characterize single precision number in Norm IEEE754. The most significant bit represent the sign of the number (one is negative and zero is positive) which is denoted by $S_{i,j}$. The following eight bits store the exponent $E_{i,j}$. The remaining 23 bits are left for the mantissa $M_{i,j}$ which can be seen from Fig.1. Then the single floating number $x_{i,j}$ can be represented as follows:

$$x_{i,j} = (-1)^{S_{i,j}} \times (1+M_{i,j}) \times (2)^{(E_{i,j}-127)} \qquad (1)$$

Some examples are displayed in Table 1.

**Table 1. Examples of Norm IEEE754 for single precision floating number.**

| Number | IEEE754-32bit |
|---|---|
| 0.3167254 | 00111110101000100010100111010101 |
| 1.2325828 | 00111111100111011100010101000110 |

## 2.3 Minimizing Additive Distortion

Almost all the modern steganographic techniques are hammer at minimizing an additive distortion function, which if well defined, is closely related to security [3]. And a quantity named cost, which is donated by $\rho_{i,j} \in [0,+\infty)$, is used to evaluate the distortion of pixel changes. To simplify our scheme, we have $\rho_{i,j}(x_{i,j}^k, x_{i,j}^k) = 0$ and $\rho_{i,j}(x_{i,j}^k, \bar{x}_{i,j}^k) = \rho_{i,j}$ in binary embedding case. The additive distortion function is in the form [6]:

$$D(X,Y) = \sum_{i=1}^{n_1}\sum_{j=1}^{n_2} \rho_{i,j} |x_{i,j} - y_{i,j}| \qquad (2)$$

Let $p_{i,j}$ be the probability of changing $x_{i,j}$ to $y_{i,j}$. Denote $m$ as the number of secret bits we want to communicate based on the binary embedding system. The optimal embedding [9] of minimizing additive distortion (1) can be simulated by allocating

$$p_{i,j} = \frac{e^{-\lambda \rho_{i,j}}}{1+e^{-\lambda \rho_{i,j}}} \qquad (3)$$

The parameter $\lambda$ can be obtained by solving

$$-\sum_{i=1}^{n_1}\sum_{j=1}^{n_2}(p_{i,j}\log_2 p_{i,j} + \bar{p}_{i,j}\log_2 \bar{p}_{i,j}) = m \qquad (4)$$

The above scenario can realize the separation of distortion function and coding system. And efficient practical codes such as STC [11] can approach the optimal embedding boundary.

## 3. STEGANOGRAPHY WITH HDR IMAGE
## 3.1 Selection on Cover

There are three established HDR image file formats in use, the HDR format, the EXR format and the TIFF format [1]. In an efficient steganography system, the information of cover image must be lossless during the communication process, which push us to choose the lossless TIFF format. Besides that, grey-scale image is usually more secure than color image especially for adaptative steganography algorithm, which can be seen from the experimental results in [12]. Taken together, the luminance channel of the original HDR image is extracted and saved as cover image with the help of libtiff library in Matlab. The image in Fig .2 is chosen as an example to show that such 32 bit grey-scale cover image won't draw any suspection from visual observation.

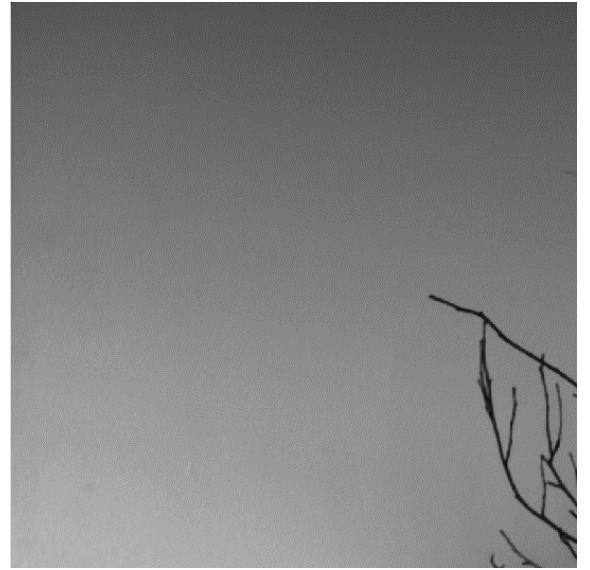

**Figure 2. An example of 32 bit grey-scale cover image.**

## 3.2 Embedding Roles

From (1) we can see that the bits used to represent mantissa are the least sensitive bits. So by employing these bits we have at most 23 bits to bear the message bits. This domain bears some similarity to the spatial domain in LDR image, so we call it mantissa spatial domain in the rest of this paper. In traditional 8-bit-plane spatial domain, we tend to change the least significant bit to minimize the embedding trace from visual observation. While in the mantissa domain, we have 23 bit planes which provide sufficient redundancy space. To simplify our design, we leave the first seven bits unchanged. Denoted the amount of the maximum embeddable bits of $x_{i,j}$ as $N_{i,j}$, the embedding rules are defined as follows:

$$N_{i,j} = \begin{cases} 16, \ x_{i,j} \geq 1 \\ E_{i,j}-111, \ x_{i,j} < 1 \end{cases} \qquad (4)$$

Single precision floating numbers have a maximum of seven decimal significant digits. The pixel value of the 32-bit cover image is always positive for which $S_{i,j} \equiv 0$. For $x_{i,j}$, flipping $x_{i,j}^k$ means that there will be $2^{(E_{i,j}-127-k)}$ change appear on itself. When $x_{i,j} \geq 1$, for $k \geq 23$ we will always have $2^{(E_{i,j}-127-k)} > 10^{-7}$. So even flipping the least significant bit, our embedding operation is effective. While for $x_{i,j} < 1$, we have $E_{i,j} < 127$. Then, $N_{i,j}$ is restricted by the significand, which results in (4). For the whole image, to simplify the design, we restrict $N_X = \min N_{i,j}$. That is, for $x_{i^*,j^*} = \min x_{i,j}$, we have $N_X = N_{i^*,j^*}$.

In LDR image steganography, the least significant bit plane is used as cover plane which is just consist of the LSB of all the pixels in the cover image. However, because of the impact of the significand, there is no such consistency in HDR image. For example, for $2^{-1} \leq x_{i,j} \leq 2^0$, the efficient LSB will be the 22th bit of the mantissa bit under the impact of exponent bias. So in our scheme, the least significant cover plane is composed by the efficient LSB of the pixels in 32-bit cover image. The rest cover planes are just in the same fashion.

## 3.3 Distortion Correction

In this paper, the design of distortion function is not what we focus on. Instead, we propose a general approach to make the well defined distortion measure for LDR image effective for HDR image. Using the distortion function of HUGO straightly on the cover image displayed in Fig.2, in Fig.3 (a) we can see that the embedding distribution is almost a random distribution. Inspired by the Local tone mapping operator [5] in HDR imaging, we defined a quantity called distortion bias and denote it by

$$\beta_{i,j} = 2^{|E_{i,j} - 127|} \quad (5)$$

After correcting each distortion $\rho_{i,j}$ by $\beta_{i,j}$, we have

$$\rho^*_{i,j} = \frac{\rho_{i,j}}{\beta_{i,j}} \quad (6)$$

The embedding probability using the corrected distortion of HUGO is displayed in Fig.3 (b).

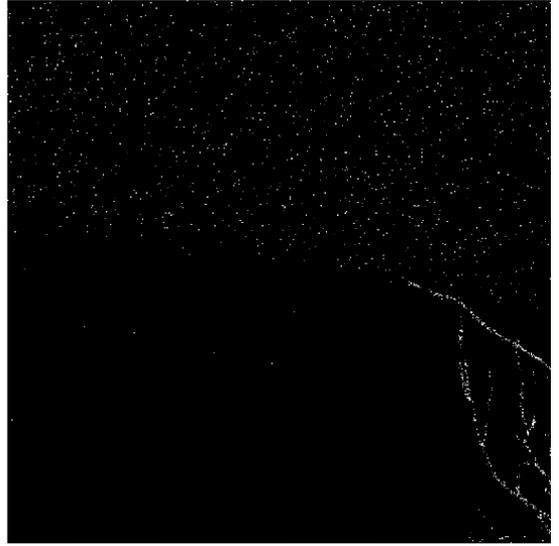

(b) using the corrected distortion function of HUGO

**Figure 3.** The embedding distribution of two cases( (a) and (b) ) for cover image(Fig.1) with payload 0.05 bit per pixel using optimal embedding. The white points in the binary images means stego pixels.

## 3.4 Practical Steganography Procedure

Assuming that we want to convey $m$ bit secret messages, the practical steganography procedure is as follows:

1. Take a HDR image and extract the luminance channel as $X$.

2. Find $x_{i^*,j^*} = \min x_{i,j}$ and calculate $N_X = N_{i^*,j^*}$.

3. Get $\bar{m} = \frac{m}{N_X}$ (To simplify the description, we assume that $\bar{m}$ is an integer. In practical, we can reallocate the secret bits to the $N_X$ cover planes.)

4. Apply a LDR distortion function on $X$ to get $\rho_{i,j}$, and get $\rho^*_{i,j}$ through $\rho^*_{i,j} = \frac{\rho_{i,j}}{\beta_{i,j}}$.

5. Extract $N_X$ cover planes $X^k (1 \leq k \leq N_X)$ and get the corresponding stego plane $Y^k$ (after embed $\bar{m}$ bit into $X^k$) with the help of STC.

6. Put the stego planes back and save $Y$ as stego image through libtiff library.

7. Convey the stego image to the receiver, and he can recover the secret message with the help of the parity check matrix of STC.

In the proposed scheme, we disperse the payload into the redundant bits to decrease the embedding changes. In practical, the relative payload $\frac{\bar{m}}{n_1 \times n_2}$ is fixed and is share to the receiver in advance. Besides that, the embedding rules and the binary

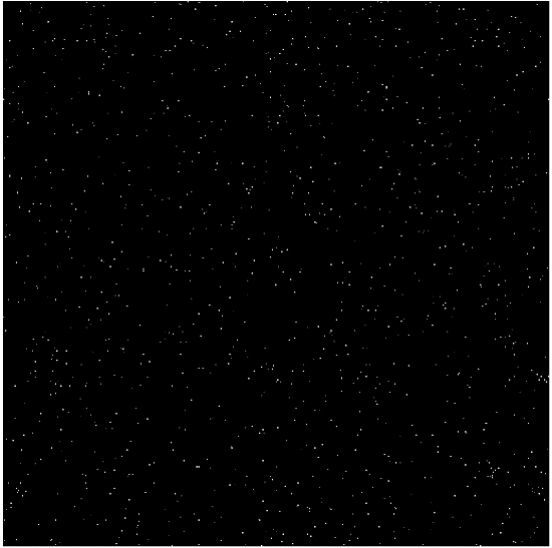

(a) using the distortion function of HUGO straightly

embedding case can make sure that there won't be any transgression problem in the receiver which guarantees the so called Non-selection channel.

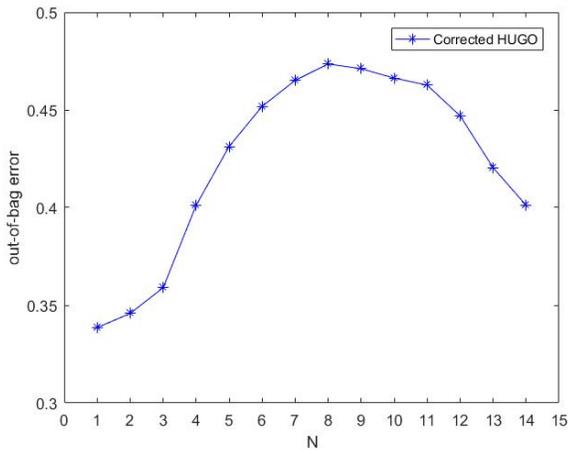

**Figure 4. The impact of embedding bits.**

## 4. EXPERIMENTS
In this section, we first assess how the embedding bits (how many bits are used to communicate secret bits per pixel) $N$ affect security without regard to the visual trace. Then we conduct two experiments on different dataset to prove the effectiveness and stability of our proposed scheme.

### 4.1 General Configuration of All Experiments
In the author's best knowledge, there is no existing steganalysis method which is specially designed for floating number. So in this paper, we make some preprocessing for the cover images and their stego versions. First, they are clamp to $[0,10^7]$ for that the minimum change of single precision floating number is $2^{-23}$. After rounding to integer, they will be turned into the SRMQ1 feature with dimension 12753.

All classifiers employed in this paper is the ensemble classifier [10] with Fisher linear discriminant as the base learner. The security of the steganography scheme is assessed using the ensemble's 'out-of-bag'(OOB) error $E_{OOB}$, which is an unbiased estimate of the minimal total testing error under equal priors, $P_E = \min_{P_{FA}} \frac{1}{2}(P_{FA} + P_{MD})$ [10].

The relative payload $\frac{\bar{m}}{n_1 \times n_2}$ is fixed to 0.05 for which the whole payload is $0.05 \times N$. And the optimal embedding simulator is used for all the algorithms employed in this paper.

### 4.2 Impact of The Embedding Bits

#### 4.2.1 Dataset
There are 106 HDR image in Fairchild dataset which is created by Fairchild in [14]. These images are split into 830 images whose size is 512*512. And the luminance channel of them is extracted and save as tiff image format with the help of libtiff library. We named it Lum-fairchild dataset.

#### 4.2.2 Experimental Results
First, we employ HUGO (binary embedding with parameters ($\lambda = 1$, $\sigma = 1$, $T = 255$) [3] to get $\rho_{i,j}$. For $1 \leq N \leq 14$, we plotting $E_{OOB}$ as a function of $N$ to see how it affect security. From Fig.4 we can see that for $1 \leq N \leq 8$, the increasing $N$ brings increasing confusion to the classifier, which is because that the pixel changes caused by $N$ is negligible when $1 \leq N \leq 8$. Keeping increasing $N$, the pixel changes tend to be big enough to be catched by the SRMQ1 feature, which results in the curve for $N \geq 9$. As displayed in Fig.4, the $E_{OOB}$ is beyond 0.45 for $6 \leq N \leq 11$, which is close to random guessing.

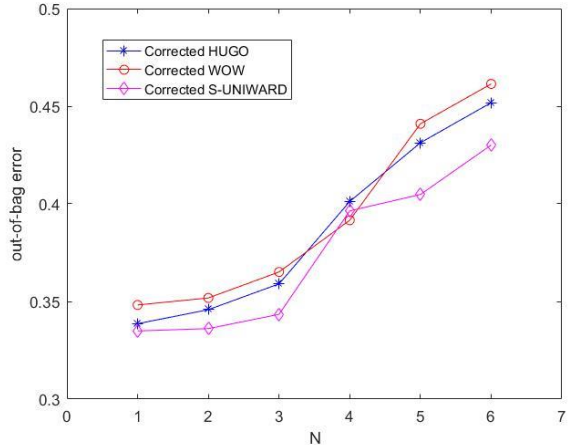

**Figure 5. The impact of distortion measure.**

After that, WOW [8] and S-UNIWARD [13] are also employed to assess the stability of the proposed scheme on different distortion measure. In experiments, we found that the whole trend of the curves is just like what Fig.4 shows. To save space, we only plot it for $1 \leq N \leq 6$. Fig.5 shows that the trend is the same for all three distortion measures.

### 4.3 Performance in Practical Case
In fact, in the proposed scheme the maximum secure embedding bits $N_{\max}$ is restricted by $N_X$ who is the minimum nonlicet $N$ which for that nonlicet $N$ will bring macroscopic embedding trace in the stego image. In other words, $N_{\max}$ is straightly restricted by the cover source. In this subsection, we will evaluate the performance of the proposed scheme in practical case based on two different dataset.

#### 4.3.1 Dataset
There are 232 HDR images in the HDRSID dataset [15]. These images are first split into about 12000 images of size 512*512. And the luminance channels of them are extracted and saved as tiff image format with the help of libtiff library. To guarantee the visual quality, we employ the images whose dynamic range is more than $2^8$. Then we restrict $N_X \geq 10$, and collect these images as the first dataset which is called Lum-HDRSID-10b. Another dataset is named Lum-HDRSID-6b for images whose $N_X \geq 6$. Then we have Lum-HDRSID-10b dataset who have 257 images and 1889-images Lum-HDRSID-6b dataset.

### 4.3.2 Experimental Results

To save space, we only employ HUGO whose configuration is just the same in subsection 4.2. For $1 \leq N_{L-H-10b} \leq 10$ and $1 \leq N_{L-H-6b} \leq 6$, we plotting $E_{OOB}$ as a function of $N$ to assess the stability of the propose scheme on different dataset. As Fig.6 shows, the trend is just the same. Besides, for $N_{L-H-10b} = 10$ (which means the real payload is 0.5bpp), we still have $E_{OOB} = 0.4883$ which prove the effectiveness and stability of our proposed scheme.

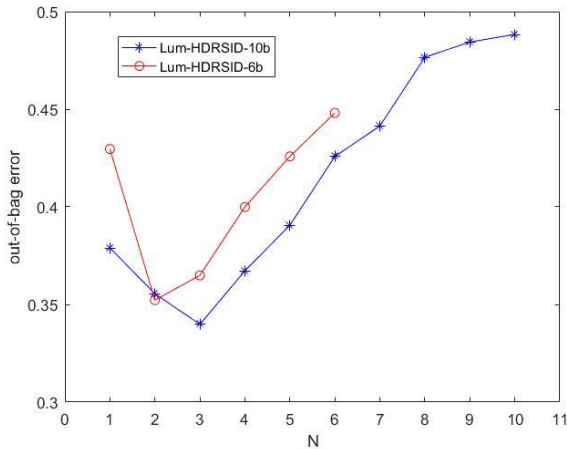

**Figure 6. The impact of dataset.**

## 5. CONCLUSIONS

In this paper, we propose a steganography scheme based on HDR image format. Norm IEEE754 and additive distortion scenario are employed to implement the embedding operation. The mantissa spatial domain is used to convey the message bits which results in less change on the cover image. Experimental results show that our scheme is effective and stable for different distortion measure and dataset.

Since the embedding rules can make good balance between high payload and security, we will try to generalize protogenetic HDR distortion measure in our future research.

## 6. ACKNOWLEDGMENTS

This work was supported by the National Science Foundation of China under Grant No. 61401072.

## Authors' background

| Your Name | Title* | Research Field | Personal website |
|---|---|---|---|
| **Wei Gao** | **Master student** | **Image steganography，HDR imaging** | |
| **Yongqing Huo** | **Associate profession** | **Image steganography，HDR imaging** | |
| **Yan Qiao** | **Master student** | **Image steganography，HDR imaging** | |